\documentclass[10pt,twocolumn,letterpaper]{article}

\usepackage{iccv}
\usepackage{times}
\usepackage{epsfig}
\usepackage{graphicx}
\usepackage{amsmath}
\usepackage{amssymb}
\usepackage{booktabs}
\usepackage{bbding}

\usepackage[breaklinks=true,bookmarks=false]{hyperref}

\iccvfinalcopy 

\def\iccvPaperID{****} 

\ificcvfinal\fi

\begin{document}

\title{An Effective Two-stage Training Paradigm Detector for Small Dataset}



\author{Zheng Wang$^{1, 2}$\and Dong Xie$^{1}$\and Hanzhi Wang$^{3}$ \and Jiang Tian$^{1}$\\
\and $\textsuperscript{1}$ AI Lab, Lenovo Research, Beijing, China\\
{\tt\small \{xiedong2, tianjiang1\}@lenovo.com}\\
$\textsuperscript{2}$ State Key Laboratory of Virtual Reality Technology and Systems, Beihang University, Beijing, China\\
{\tt\small zheng.wang@buaa.edu.cn}\\
$\textsuperscript{3}$ Department of Computer Science, Yale University, New Haven, Connecticut, USA\\
{\tt\small daniel.wang.dhw33@yale.edu}
}

\maketitle
\ificcvfinal\thispagestyle{empty}\fi

\begin{abstract}
Learning from the limited amount of labeled data to the pre-train model has always been viewed as a challenging task. In this report, an effective and robust solution, the two-stage training paradigm YOLOv8 detector (TP-YOLOv8), is designed for the object detection track in VIPriors Challenge 2023. First, the backbone of YOLOv8 is pre-trained as the encoder using the masked image modeling technique. Then the detector is fine-tuned with elaborate augmentations. During the test stage, test-time augmentation (TTA) is used to enhance each model, and weighted box fusion (WBF) is implemented to further boost the performance. With the well-designed structure, our approach has achieved $30.4 \%$ average precision from $0.50$ to $0.95$ on the DelftBikes test set, ranking 4th on the leaderboard.
\end{abstract}

\section{Introduction}
Object detection is a versatile technology that finds extensive use in numerous fields. However, the training of detectors require a large amount of annotated data, which is labor-intensive and expensive. In order to improve data efficiency, the 4th Visual Inductive Priors for Data-Efficient Deep Learning Workshop(VIPriors) is introduced as an ICCV 2023 workshop. In particular, the use of model weights pre-trained on large-scale datasets is not allowed. Contestants should train a detector on the given dataset from scratch. The dataset used in the object detection track is DelftBikes\cite{delftBikes}, which contains 10,000 bike pictures, with 22 densely annotated parts per image, where some parts may be missing. The dataset is split into 8,000 training and 2,000 testing images. 

\begin{figure}
    \begin{center}
        \begin{tabular}{c} 
            \includegraphics[height=7.5cm]{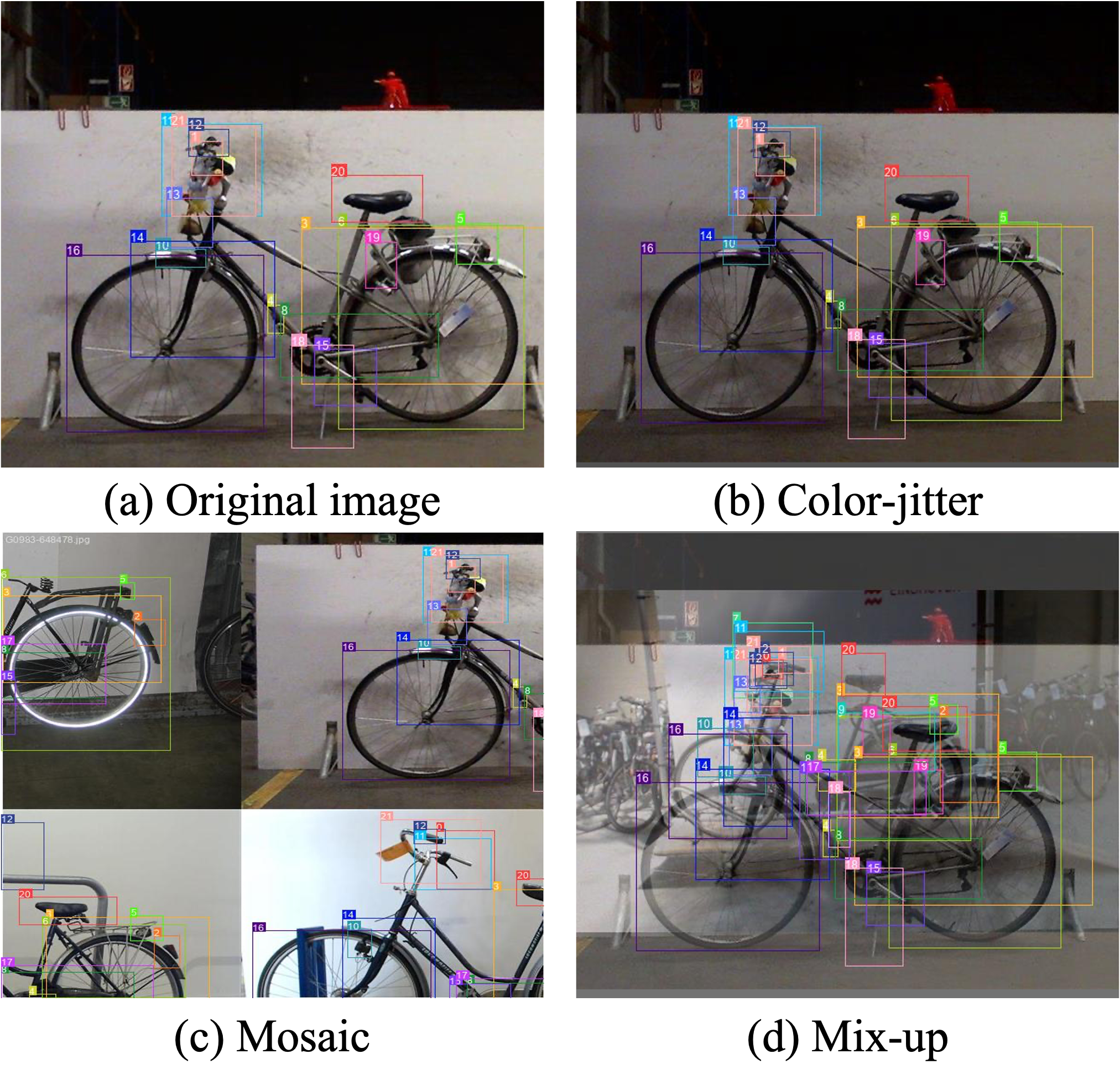}
        \end{tabular}
    \end{center}
  \caption{Visual examples of our data augmentation methods. (a) The original image and ground truth boxes. (b) The image after color jittering. (c) The image is cut and pieced together from segments of three additional pictures. (d) The image is mixed with an additional picture to a certain proportion.}
  \label{fig:augmentation}
\end{figure}

\begin{figure*}
    \begin{center}
        \begin{tabular}{c} 
            \includegraphics[height=6cm]{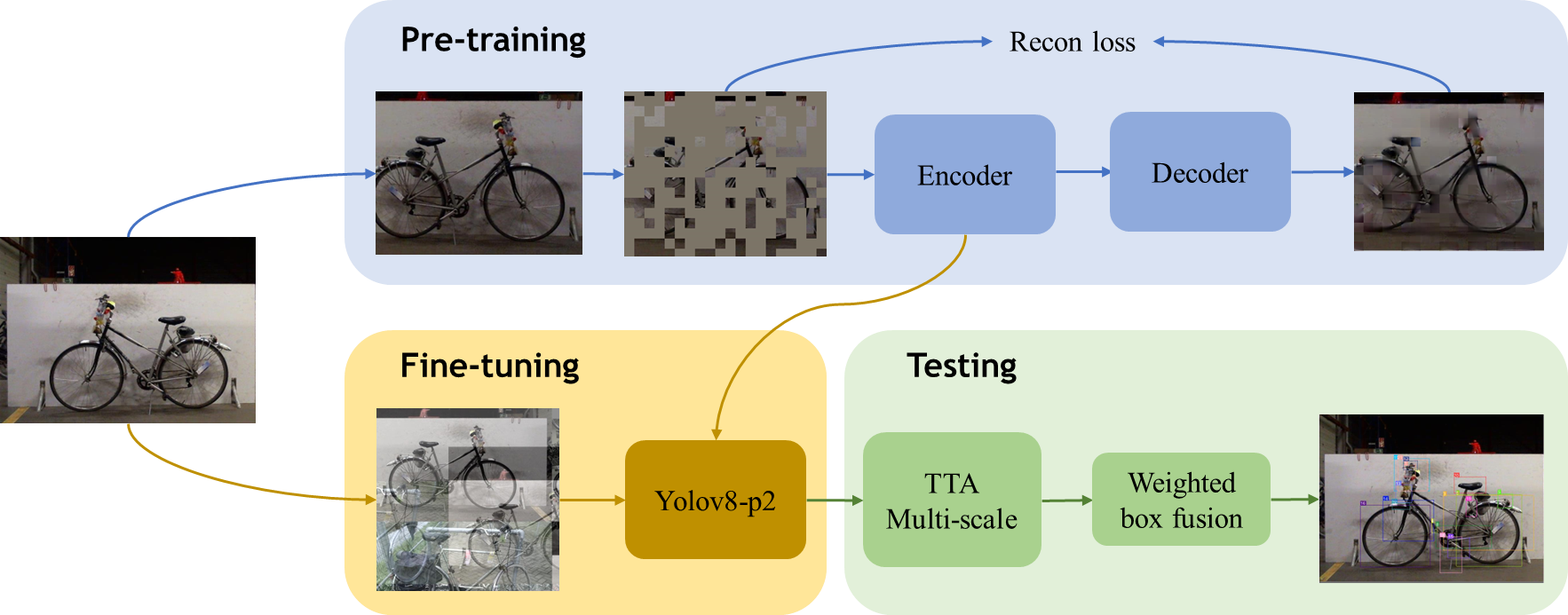}
        \end{tabular}
    \end{center}
  \caption{Our two-stage training paradigm TP-YOLOv8. To perceive relative position distribution between bicycle parts, we fisrt pre-train the backbone with the MIM technique in the blue part. Then the pre-trained weight is loaded for fine-tuning on the detection task using elaborate augmentations within the yellow background. In the testing stage, TTA and model ensembling methods are implemented to further boost the detection performance depicted in the green area.}
  \label{fig:structure}
\end{figure*}

We experiment with various models. The challenge does, however, limit the usage of extra data and pre-trained weights. Finally, the lightweight detector YOLOv8\cite{yolov8} is selected as our baseline. In order to address the data deficiency problem, a two-stage training method is designed in which the backbone of the detector is pretrained with unsupervised masked image modeling(MIM)\cite{spark, bert} method and then the detector is fine-tuned with elaborate data augmentations. In the pretraining stage, an MIM technique named SparK is utilized to help the detector learn better features by integrating prior knowledge about positions into the detector. Then we load the pretrained encoder as the backbone and finetune it with elaborate augmentation method such as Color Jitter, Mosaic\cite{yolov4} and Mix-up\cite{mixup}. The visual examples of our data augmentation methods are shown in Figure \ref{fig:augmentation}. In the testing stage, TTA and weighted box fusion\cite{wbf} are implemented to further boost the detection performance. Our approach achieves $30.4 \%$ AP on the DelftBikes test set, ranking 4th on the leaderboard.

The implementation details of our method are introduced in the following sections.


\section{Methodology}

For a given image, we first use MIM pre-trained encoder to extract image features. Then we send the extracted features to the YOLOv8 to detect the objects.
In this section we first illustrate the pre-training method named SparK\cite{spark} in Sec.2.1. Then augmentations used in fine-tuning stage is introduced in Sec.2.2. And the TTA and model ensembling strategy is presented in Sec.2.3. The framework s depicted in Figure \ref{fig:structure}.

\subsection{Pre-training Stage}
The challenge dataset DelftBikes is relatively small. Experiments show that direct training on the training set leads to overfitting. In order to avoid this issue and fully utilize the positional prior knowledge, we follow the popular masked image modeling approach to pre-train the backbone of YOLOv8. 

The masked image modeling initially extending the success of BERT from transformers to vision transformers, is largely increasing the performance of ViTs. However, the masked approach is difficult to be applied for hierarchical convolutional models. SparK treats unmasked patches as sparse voxels and uses sparse convolution to encode them. SparK also employs a hierarchical decoder to make full use of the advantage of the convolution net’s hierarchy, which makes masked modeling well-suited for any convolution net, and brings a performance leap on downstream tasks.

To adapt convolution to irregular masked input, visible patches are gathered into a sparse image and encoded by sparse convolution. To pre-train a hierarchical encoder, SparK engages a UNet-style\ architecture\cite{u-net} to decode multi-scale sparse feature maps, where all empty positions are filled with mask embedding. SparK's reconstruction target is per-patch normalized pixels with an $L^2$-loss.

\subsection{Fine-tuning Stage} 
After pre-training, we dessert the decoder in the SparK and use the encoder as the backbone of the detector. We also follow the augmentation methods in the Yolo series, such as mosaic and mix-up to improve the generalization of the detector.

The size of some object categories in the DelftBikes dataset is relatively small. With multi-scale training and testing and tweaking the architecture to detect on a larger feature map, i.e., the YOLOv8-p2, our model achieves better performance, especially on small objects. The ablation study is displayed in Sec.4.2.

\subsection{Testing Stage}
Before testing, we augment the images in the test set with methods like RandomFlip and Resize, which brings more scale-wise diversity. After acquiring all the detection results, we sort them by AP on the validating set and implement weighted box fusion on models with the best performance.

\begin{table*}[h]
\centering
\caption{The 2023 VIPriors object detection challenge final leaderboard.}
\label{tab:leaderboard}
\begin{tabular}{cccccccc}
    \toprule
    Rank & User & AP@0.50:0.95& AP@0.50 & AP@0.75 & AP@(small) & AP@(medium) & AP@(large) \\
    \midrule
    1 & w12 & 0.35 (1) & 0.66 & 0.29 & 0.15 & 0.31 & 0.25 \\
    2 & GroundTruth & 0.33 (2) & 0.69 & 0.28 & 0.16 & 0.27 & 0.27 \\
    3 & HHHa & 0.31 (3) & 0.64 & 0.25 & 0.12 & 0.28 & 0.24 \\
    \textbf{4} & \textbf{wokots} & \textbf{0.30 (4)} & \textbf{0.63} & \textbf{0.25} & \textbf{0.11} & \textbf{0.27} & \textbf{0.24} \\
    \bottomrule
\end{tabular}
\end{table*}

\begin{table*}[h]
\centering
\caption{The ablation experiments of our method.}
\label{tab:ablation}
\begin{tabular}{cccccccc}
    \toprule  
    Model& mosaic&	mixup& pre-train& mask ratio&		AP@0.50:0.95(\%) & AP@0.50(\%) & AP@0.75(\%) \\
    \midrule  
    Yolov8-p2& \XSolidBrush& \XSolidBrush& \XSolidBrush& -& 30.0&	62.4&	24.2 \\
	-& \CheckmarkBold& \XSolidBrush& \XSolidBrush& -& 30.1&	62.5&	24.4 \\
	-& \CheckmarkBold& \CheckmarkBold& \XSolidBrush& -& 30.1&	62.4&	24.7 \\
	-& \CheckmarkBold& \CheckmarkBold& \CheckmarkBold&	cut 0.75&		30.2& 	62.5& 	24.6 \\
	-& \CheckmarkBold& \CheckmarkBold& \CheckmarkBold&	cut 0.60&		30.1& 	62.3& 	25.0 \\
	-& \CheckmarkBold& \CheckmarkBold& \CheckmarkBold&	whole 0.75&		30.2& 	62.8& 	24.8 \\
	-& \CheckmarkBold& \CheckmarkBold& \CheckmarkBold&	whole 0.60&		30.3& 	62.6& 	24.9 \\
    Ensemble& -& -& -& -& 31.4& 	64.5& 	26.3 \\ 
 
    \bottomrule 
\end{tabular}
\end{table*}

\begin{table}[h]
\centering
\caption{Experimental results on DelftBikes validation set.}
\label{tab:exp_all_model}
\begin{tabular}{cccc}
    \toprule  
    Method & AP@0.50:0.95& AP@0.50& AP@0.75 \\
    \midrule  
    Yolov8M&	29.5& 	62.4& 	23.0 \\
    Yolov8L&	29.6& 	62.4& 	24.1 \\
    Yolov5M&	29.6& 	61.9& 	24.1 \\
    Yolov5L&	29.5& 	62.0& 	23.8 \\
    Yolov5L-p2&	30.0& 	62.7& 	24.4 \\
    Yolov8M-p2&	30.1& 	62.5& 	24.7 \\
    Yolov8L-p2&	30.2& 	62.6& 	24.9 \\
    \bottomrule 
\end{tabular}
\end{table}

\section{Experiment}
\subsection{Implement Detail}
All of our experiments are conducted on 1 V100 GPU. In the pre-training stage, the backbone of YOLOv8 is pre-trained by reconstructing images from multi-scale encoded features with a batch size of 64 and a mask ratio of 60 for 1,000 epochs. After pre-training, we load the pre-trained weight and fine-tune the detector with annotated images. During the fine-tuning stage, we set the learning rate of the backbone to 0.001 and that of other parts of the detector to 0.01. The detector YOLOv8 is trained for 100 epochs with several different settings. In the testing stage, models with different hyperparameters are evaluated on the validation set and we take the best-performing models for weighted box fusion. In the end, we take the top 30 models for model ensembling, improving the mAP by $1.2\%$.
\subsection{Result Analysis}
In the 2023 VIPriors object detection challenge, our team, wokots, showcased its expertise and dedication by securing a commendable 4th position on the Table \ref{tab:leaderboard}. Analyzing our scores in detail, we achieved an Average Precision (AP) score of 0.30 at an Intersection over Union (IoU) threshold ranging from 0.50 to 0.95. This score, while being competitive, was just a hair's breadth behind the 3rd place team, HHHa, who managed an AP of 0.31. At an IoU threshold of 0.50, our model demonstrated an impressive AP score of 0.63, indicating a strong ability to detect objects with a significant overlap with ground truth. For a more stringent IoU threshold of 0.75, our AP score stood at 0.25, reflecting our model's capability to maintain precision even under stricter evaluation criteria. Diving deeper into size-specific evaluations, our model exhibited a nuanced understanding of object scales. For small objects, which often pose challenges due to their limited visibility and intricate details, we achieved an AP of 0.11. For medium-sized objects, our AP was a robust 0.27, and for large objects, which require the model to capture broader contexts, our AP was a solid 0.24.

We evaluated different Yolo variants on the validation set under the same set of hyperparameters, including Yolov8M/L and Yolov5M/L and the scale-enhanced versions of these models. As shown in Table \ref{tab:exp_all_model}, These scale-enhanced configurations, especially Yolov8L-p2 consistently outperformed their base counterparts across all metrics. Finally we choose Yolov8L-p2 as our baseline model.

In our ablation study, we evaluated the effectiveness of our augmentation strategies and the MIM pre-training. We also experiment with a bigger mask ratio and pre-training on images with the bikes cut out. The ablation experiments are illustrated in Table \ref{tab:ablation}.

\section{Conclusion}
In our method, a two-stage training paradigm named TP-YOLOv8 is proposed. The usage of masked image modeling unsupervised pre-training strategy injects prior knowledge of relative positions of bicycle parts into the model, greatly improving the performance of the models. Besides, test-time augmentation and weightd box fusion are implemented to further boost the performance. These combined methodologies and techniques have proven their efficacy, as evidenced by our commendable achievement of securing the 4th position in the VIPriors object detection challenge. This accomplishment underscores the potential and effectiveness of our proposed method in the realm of object detection.

{\small
\bibliographystyle{ieee_fullname}
\bibliography{VIPriors2023/report}
}

\end{document}